\documentclass[a4paper, 10pt, twocolumn, twoside]{article}

\usepackage{ISARC}
\usepackage{xcolor}

\begin{document}

\linespread{0.5}

\title{Event-based object detection under adverse lighting and motion: lessons for construction sites}

\author{
    Zhuoran Feng$^{1}$, Xiaoshan Zhou$^{2,*}$ \\[0.5em]
    \small $^{1}$School of Computer Science, Faculty of Engineering, The University of Sydney, NSW, Australia \\[0.3em]
    \small $^{2}$School of Project Management, Faculty of Engineering, The University of Sydney, NSW, Australia \\[0.3em]
    \small $^{*}$Corresponding author: \texttt{xiaoshan.zhou@sydney.edu.au}
}

\maketitle
\thispagestyle{fancy}
\pagestyle{fancy}

\begin{abstract}
Vision-based monitoring on construction sites fails at night, at dawn and dusk, in tunnels and under floodlights, when conventional RGB cameras under- or over-expose. Event cameras record per-pixel brightness changes with microsecond latency and over 120 dB of dynamic range, but their benefit for site-relevant object detection has not been quantified on real data. We compare three recent event-based detectors (SMamba, SATE and EvRT-DETR) with an RGB detector trained on the same data, using the pixel-aligned PEOD dataset. On low-light frames (70\% of the test set) the best event detector reaches 27.2 AP against 21.3 AP for RGB, and a training-free late fusion of both sensors is the best option in every lighting condition (33.3 AP). An object-level analysis of 53,091 objects shows that brightness has little effect on event detectors, whereas fast motion matters: above 20 pixels per frame RGB finds 11--19\% of medium-size objects and EvRT-DETR 39--60\%. Activation patching locates the damage caused by losing events at the 1/16-resolution features, with small objects depending on the finest scale, and shows that event detectors fail when few signal events meet added noise. A model-agnostic pruning study keeps about 95\% of accuracy with 60\% of the input but halves small-object recall. On rendered construction sites, used zero-shot, dimming the floodlights drops RGB to near zero while the event detector keeps most of its AP$_{50}$. We close with camera-placement guidance for construction sites.
\end{abstract}

\begin{keywords}
event camera, object detection, construction safety, low-light vision, token pruning
\end{keywords}

\section{Introduction}

Camera-based detection of workers, vehicles and plant underpins construction-site safety monitoring, progress tracking and the perception of autonomous machinery~\cite{fang2020cvsafety,paneru2021cv}. These systems almost always use RGB cameras, and they inherit the RGB camera's weakness: a fixed exposure and a dynamic range of about 60 dB. Construction work does not stop when the light is poor. Road and rail works are moved to night shifts to avoid traffic, concrete pours start before dawn, and tunnels and basements are lit only by work lamps. These are also the conditions in which workers are most at risk: an analysis of fatal crashes in highway work zones found night-time work zones more hazardous than daytime ones~\cite{arditi2007night}.

\begin{figure*}[!htb]
\centering
\includegraphics[width=\textwidth]{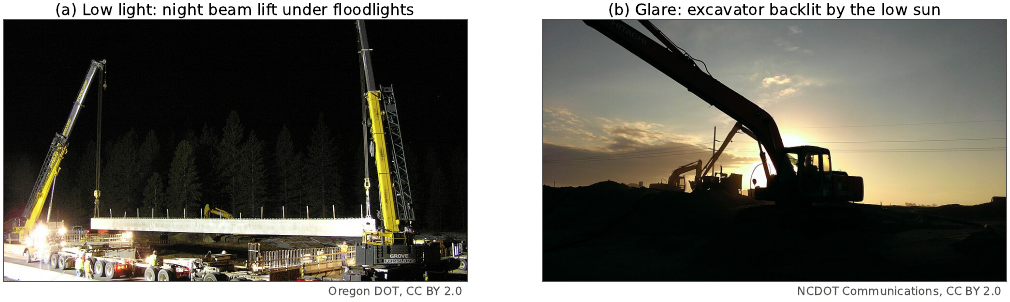}
\caption{Two lighting conditions that defeat RGB cameras on real construction sites. (a) Night work: a bridge-beam lift lit by floodlights, with most of the scene close to black. (b) Glare: an excavator backlit by the low sun, saturated sky and underexposed foreground in the same frame. Photographs: (a) Oregon Department of Transportation, (b) NCDOT Communications, both CC BY 2.0, via Wikimedia Commons.}
\label{fig:photos}
\end{figure*}

Two failure modes follow from the limited dynamic range (Figure~\ref{fig:photos}). In low light the sensor collects too few photons, so the camera raises its gain and lengthens its exposure; the image becomes noisy, moving workers and plant smear, and dark objects merge with the background (Figure~\ref{fig:photos}a). In high-contrast scenes, such as a low sun behind an excavator, floodlights pointing at the camera or a vehicle leaving a dark tunnel, the bright part of the image saturates while the foreground stays underexposed (Figure~\ref{fig:photos}b), and auto-exposure needs time to adapt. Studies of construction vision under difficult lighting have so far worked on the RGB image itself: an improved YOLOX with contrast enhancement detects small personal protective equipment in tunnels~\cite{wang2023yolox}, a YOLOv8 tracker with self-calibrated illumination follows workers and machinery in tunnels~\cite{li2026tunnel}, and illumination enhancement supports worker pose estimation at night~\cite{chen2022uire}. Others add a thermal camera and fuse it with RGB in dust and fog~\cite{rgbtmt2025}. Enhancement cannot recover detail that the sensor never recorded, in saturated pixels or under motion blur, and these works report accuracy on their own datasets without separating low light from overexposure or from motion.

Event cameras follow a different sensing principle. Each pixel independently reports a change in log brightness as soon as it exceeds a contrast threshold, giving microsecond temporal resolution, no motion blur from a fixed exposure and a dynamic range above 120 dB~\cite{gallego2022event}. Because the sensor responds to relative change, a dark scene and a bright scene produce similar signals as long as something moves. Event-based object detection has advanced quickly on automotive benchmarks, from recurrent vision transformers~\cite{gehrig2023rvt} to sparse state-space models~\cite{smamba2025}, scale-aware temporal encoders~\cite{sate2025} and adapted image detectors~\cite{evrtdetr2025}. In civil engineering, event cameras have been used to record cracks and spalling for infrastructure inspection in poor lighting~\cite{gamage2025evcivil}. For construction, one workshop study trained YOLOv8 and DETR on event frames from a simulated construction site seen by a mobile robot~\cite{guaman2024event}, without an RGB baseline and under a single lighting condition. We found no study that evaluates event-based object detection on real recordings under the lighting and motion conditions of construction sites, or that compares it with an RGB detector trained on the same data.

What prevents an informed decision on whether to install event cameras on a site is not the lack of a sensor but the lack of evidence on how event detection behaves under the two factors that define site conditions: lighting and motion. Five questions remain open, and each motivates one part of this study. First, published comparisons report events and RGB on aggregate benchmarks; the PEOD benchmark found that the best event-only detector beats every fusion model on its illumination-challenge subset~\cite{peod2026}, but did not split this subset into low light and overexposure. A site manager needs to know, per lighting condition, whether events beat an RGB detector trained on the same data. Second, the two sensors are expected to fail in different conditions, which suggests combining them; learned fusion needs paired training data that sites rarely have, so it matters whether a training-free fusion of the two detectors' outputs already beats both. Third, an event camera responds to brightness change, which depends on motion as well as on illumination. In darkness, however, noise events make up a larger share of the event stream~\cite{gallego2022event}, so it is not obvious whether brightness or motion governs event detection, and the two have not been separated on real data. Fourth, event detectors differ widely in accuracy, and choosing one for a site requires knowing whether the differences come from pre-training data or from architecture, and which internal features carry the damage when events are lost in the dark. Fifth, event data are sparse, and on-site devices have limited compute. Vision transformers can drop uninformative tokens with little loss~\cite{rao2021dynamicvit,liang2022evit,bolya2023tome}, and some event detectors select tokens internally~\cite{smamba2025}, but pruning has been studied within single architectures, so it is unknown whether redundancy is a property of the event data and what is lost when it is removed.

We answer these questions with a controlled benchmark on PEOD~\cite{peod2026}, which provides pixel-aligned event and RGB streams with illumination labels, and test the conclusions zero-shot on three rendered construction sites. Our contributions are:

\begin{enumerate}[noitemsep]
\item \textbf{A fair RGB-vs-event comparison by lighting condition.} With the RGB detector trained on PEOD, the best event detector leads by 5.9 AP in low light (27.2 vs 21.3), while overexposed frames are a tie.
\item \textbf{Training-free RGB--event fusion} that beats both sensors in every lighting condition (33.3 AP vs 32.2 for events and 27.1 for RGB).
\item \textbf{Brightness versus motion.} An object-level analysis separates object size, apparent speed and frame brightness: for event detectors brightness has a small effect and speed a larger one, and the event advantage over RGB grows with speed. Darkening identical scenes confirms that RGB collapses beyond its gain limit while event detection degrades gradually.
\item \textbf{Inside the detectors.} Controlled training runs attribute the gap between event detectors to pre-training (about 4 AP) and architecture (about 6 AP); activation patching, noise injection, memory ablation and probes show where losing events hurts, that small and large objects rely on different scales, and that the signal-to-noise ratio rather than brightness breaks event detection.
\item \textbf{Pruning and deployment.} Keeping 60\% of input regions retains about 95\% of AP in three architectures, but small-object recall halves even when their regions are kept; together with the simulated sites, this yields camera-placement guidance for construction sites.
\end{enumerate}

Section~2 describes the data, detectors and analyses, Section~3 reports the results, and Sections~4 and~5 discuss implications and conclusions.

\section{Method}

\subsection{Dataset and lighting subsets}

PEOD~\cite{peod2026} was recorded with a Prophesee EVK4 HD event camera (1280$\times$720) and a Hikvision MV-CS050-10UC industrial RGB camera that share one optical path through a 50:50 non-polarising beam splitter, with identical 25 mm lenses and a fixed aperture. A signal generator triggers both cameras, and stereo rectification aligns the two streams pixel by pixel, so a box drawn in one stream is valid in the other. The rig was mounted behind the windscreen of a car and recorded more than 130 driving sequences of up to 90 s on urban and suburban roads, intersections, tunnels and highways between 04:00 and 24:00, from dawn to night. Boxes are annotated at 30 Hz for six classes (car, person, bus, truck, 2-wheeler and 3-wheeler); in poor lighting the annotators drew them on grey images reconstructed from events, so the labels do not depend on what the RGB camera saw. In total PEOD contains over 72,000 annotated frames and 340,000 boxes.

We use the 97 training and 24 test sequences of the official split, converted to the RVT format. The test split has 11,659 labelled frames with 53,091 boxes; the RGB detector is trained on the 59,628 training RGB frames. Event detectors receive, for each labelled frame, the events of the preceding 33 ms as a 20-channel histogram (10 time bins $\times$ 2 polarities, RVT format).

\begin{figure*}[!htb]
\centering
\includegraphics[width=\textwidth]{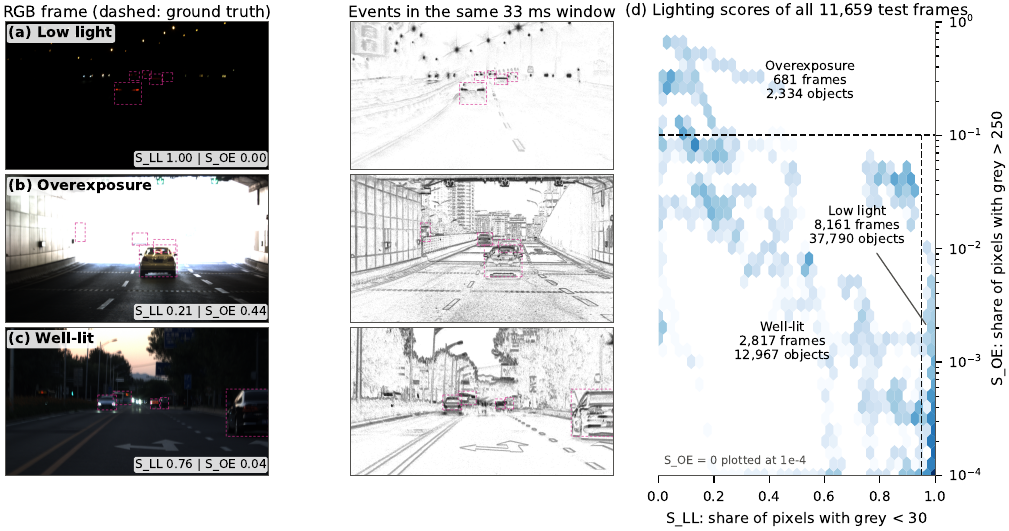}
\caption{PEOD test data. (a)--(c) One frame per lighting subset: RGB (left) and the events of the same 33 ms window (right; darker = more events), with ground-truth boxes dashed. (d) Lighting scores of all test frames; dashed lines are the subset thresholds (S\_OE $\geq$ 0.10: overexposure; otherwise S\_LL $\geq$ 0.95: low light).}
\label{fig:dataset}
\end{figure*}

Following PEOD, we describe the lighting of each frame with two scores computed on the grey-level RGB image: the low-light score S\_LL is the share of pixels darker than 30 and the overexposure score S\_OE is the share brighter than 250 (8-bit scale). A frame with S\_OE $\geq$ 0.10 is \emph{overexposed} (681 frames); of the rest, a frame with S\_LL $\geq$ 0.95 is \emph{low light} (8,161); the remaining 2,817 frames are \emph{well-lit}. Figure~\ref{fig:dataset} shows one frame of each subset in both modalities and the distribution of the two scores. In the low-light example almost nothing is visible in RGB, whereas the event stream still outlines the vehicles; in the overexposed example the tunnel exit saturates the RGB image while the event stream keeps the structure beyond it. Low light covers 70\% of the test frames and 37,790 of the objects, so the PEOD test set is dominated by night driving.

\subsection{Detectors}

We test three recent event-based detectors that represent the main architectural families, and an RGB detector of similar capacity (Table~\ref{tab:det}).

\textbf{SMamba}~\cite{smamba2025} replaces the attention blocks of a hierarchical event backbone with Mamba state-space layers, whose cost grows linearly with the number of tokens. Its key idea is to process only informative tokens: a spatio-temporal continuity assessment (STCA) scores each token from the temporal continuity of its events, which separates object activity from isolated noise, and discards low-scoring tokens. An information-prioritised local scan orders the remaining tokens so that tokens of the same object stay close in the sequence, and a global channel interaction (GCI) module restores global context along the channel dimension. Because the GCI layers depend on the input size, weights pre-trained on 1Mpx transfer only at half resolution.

\textbf{SATE}~\cite{sate2025} revisits where temporal memory sits in an event detector. Most detectors add recurrent layers only to the coarsest, high-level features; SATE adds them at the early, high-resolution scales as well, through a multi-branch downsampling design in which each scale has its own spatio-temporal encoding. Its decoupled deformable-enhanced recurrent layers separate motion estimation from feature fusion, and a feature pyramid combines the scales. Early temporal encoding is meant to preserve small, fast objects, which matters for distant workers on a site.

\textbf{EvRT-DETR}~\cite{evrtdetr2025} takes the opposite route and adapts an image detector rather than designing an event-specific one. The events are stacked into an image-like tensor and fed to RT-DETR~\cite{zhao2024rtdetr}, a real-time detection transformer with a ResNet-50 backbone, a hybrid encoder and a query-based decoder that needs no non-maximum suppression. Training has two stages: the detector is first trained on single event frames, then frozen while lightweight recurrent (ConvLSTM) modules are inserted into the encoder's latent space at several scales to add temporal memory. This design lets the detector reuse large-scale image pre-training and the 1Mpx event pre-training released by the authors.

\textbf{RGB baseline.} The RGB detector is YOLOv8x~\cite{yolov8}, a widely used one-stage detector, run either zero-shot with COCO weights or fine-tuned from them at 1280 px on the 59,628 PEOD training RGB frames, with 10\% of the training sequences held out for validation; training stopped at a 5 h time limit after 15 epochs.

\begin{table}[!htb]
\centering
\caption{Detectors compared (res.: input resolution).}
\label{tab:det}
\small\setlength{\tabcolsep}{3pt}
\begin{tabular}{llcl}
\hline
\textbf{Model} & \textbf{Pre-training} & \textbf{Res.} & \textbf{Training on PEOD} \\ \hline
SMamba & none & full & 120k steps \\
SMamba + 1Mpx & 1Mpx & half & 20k steps \\
SATE & none & full & 60k steps, bf16 \\
EvRT-DETR & 1Mpx (Gen4) & half & 2 stages, 30 epochs \\
RGB zero-shot & COCO & full & none \\
RGB trained & COCO & full & 15 epochs \\
\hline
\end{tabular}
\end{table}

\subsection{Evaluation}

We report COCO AP, AP$_{50}$ and AP by object size on the same frames and ground truth for every model, with detections at confidence $\geq$ 0.1 unless noted. For the zero-shot RGB detector, COCO classes map to PEOD classes (bicycle and motorcycle $\rightarrow$ 2-wheeler) and the 3-wheeler class is missing. For object-level comparisons, a ground-truth box counts as \emph{found} if a same-class detection with confidence $\geq$ 0.3 overlaps it with IoU $\geq$ 0.5. Run-to-run noise is about $\pm$0.2 AP. To account for variation between scenes, we also give 95\% confidence intervals from a bootstrap over the 24 test sequences (100 resamples, paired across models). All training and inference ran on a single NVIDIA H200 GPU.

\subsection{Late RGB--event fusion}

Because PEOD's streams are pixel-aligned, the boxes of an RGB detector and of an event detector live in the same image coordinates and can be merged without calibration. We run trained YOLOv8x on the RGB frame and EvRT-DETR on the 33 ms event histogram that ends at the same timestamp; EvRT-DETR's half-resolution boxes are scaled back to full resolution. For each labelled frame we take the union of the two detection sets and apply class-wise non-maximum suppression (NMS) at IoU 0.6: when both sensors detect the same object, the box with the higher confidence survives, and an object seen by only one sensor keeps that sensor's box. Confidences are used as given and no parameter is tuned on the test set. Fusion can thus recover objects that one sensor misses, but it cannot remove a false positive produced by either sensor. As variants, we multiply the event detector's confidences by a weight per lighting subset, and we calibrate each detector's confidences with Platt scaling~\cite{platt1999} (a logistic fit of whether a detection is correct, IoU $\geq$ 0.5, on its log-odds), globally or per lighting subset. Weights and calibrators are fitted by 4-fold cross-validation over sequences, so no frame is scored with parameters fitted on its own sequence.

\subsection{Object-level analysis of brightness and motion}

AP mixes many objects of different size, speed and brightness. To separate these factors we analyse every test object individually (53,091). For each object we record its box area, the brightness of its frame (mean grey level of the RGB image) and its apparent speed. Speed is measured from the labels rather than simulated: we match the box to the same object in the previous labelled frame, by track identity where available or otherwise as the nearest box of the same class within $\max(64, 0.5\sqrt{wh})$ pixels, and divide the displacement of the box centre by the time between the frames (at most 100 ms), normalised to pixels per 33 ms. Speed is known for 52,211 objects (98\%); its median is 4.5 px per frame and 6.8\% of objects move faster than 20 px per frame. Apparent speed combines object motion and the motion of the car-mounted camera. For each detector we fit a logistic regression of \emph{found} on standardised log speed, log brightness and log box area, so the coefficients are comparable effects of a one-standard-deviation change, and we report detection rates of medium-size objects (32$^2$--96$^2$ px) in five speed bins with Wilson 95\% intervals. Overexposed frames are excluded because they are too few for this breakdown.

\subsection{Model-agnostic token pruning}

To compare pruning across architectures, we prune the input rather than internal tokens. Each event frame is divided into tiles (16$\times$16 px at full resolution, 8$\times$8 px at half); we keep the fraction k of tiles with the most events and set the rest to zero, for k $\in$ \{1.0, 0.6, 0.4, 0.2, 0.1\}. The same hook runs unchanged for all three detectors. This measures spatial redundancy of the input; it does not by itself reduce computation. For reference, we also report SMamba's internal STCA token selection at similar keep ratios.

\subsection{Interpretability probes}

We use four tools on the trained detectors, without retraining:

\begin{itemize}[noitemsep]
\item \textbf{Activation patching}~\cite{meng2022rome}\textbf{.} Input event counts are thinned at random to a share \emph{k} of the original (and optionally noise events are added), mimicking a darker scene. A parallel run on the full input supplies the features of one scale (1/8, 1/16 or 1/32), and the AP recovered by restoring that scale measures how much of the damage it carries.
\item \textbf{Memory ablation.} EvRT-DETR's ConvLSTM memory is reset at every frame, so each frame is processed on its own.
\item \textbf{Linear probes}~\cite{alain2016probes}\textbf{.} Logistic regressions decode low light versus well-lit from spatially pooled SMamba features at each stage, evaluated leave-one-sequence-out and compared with a probe on the event count alone.
\item \textbf{Activation steering}~\cite{turner2023steering}\textbf{.} The mean feature difference between well-lit and low-light frames is added to the features of low-light frames.
\item \textbf{Denoising.} After event loss and noise, a background-activity filter~\cite{delbruck2008baf} keeps an event-histogram cell only if at least $f$ (1 or 2) of its neighbours in a 3$\times$3-pixel, 3-time-bin window are active.
\end{itemize}

\subsection{Controlled darkening}
 
Low-light and well-lit PEOD frames come from different scenes. To compare the sensors on identical scenes, we darken the six brighter test sequences by 0 to 7 stops (each stop halves the light), calibrated on PEOD itself. Measured on every 10th labelled frame, low-light frames are 6.7 stops darker than well-lit ones (median linear brightness) and carry 38\% of the clustered events, while the density of isolated, noise-like events does not increase. For RGB, each frame is linearised, its photon count is scaled and resampled with Poisson shot noise and 5 e$^-$ read noise (full well 10,000 e$^-$), amplified by auto-gain up to 16$\times$ and re-encoded to 8-bit sRGB before running trained YOLOv8x. For events, the input is thinned to the measured share, interpolated in log brightness (100\% at 0 stops, 38\% at 6.7 stops), either alone or with noise added so that the absolute noise density stays at its daytime level.

\subsection{Simulated construction sites}

\begin{figure*}[!htb]
\centering
\includegraphics[width=\textwidth]{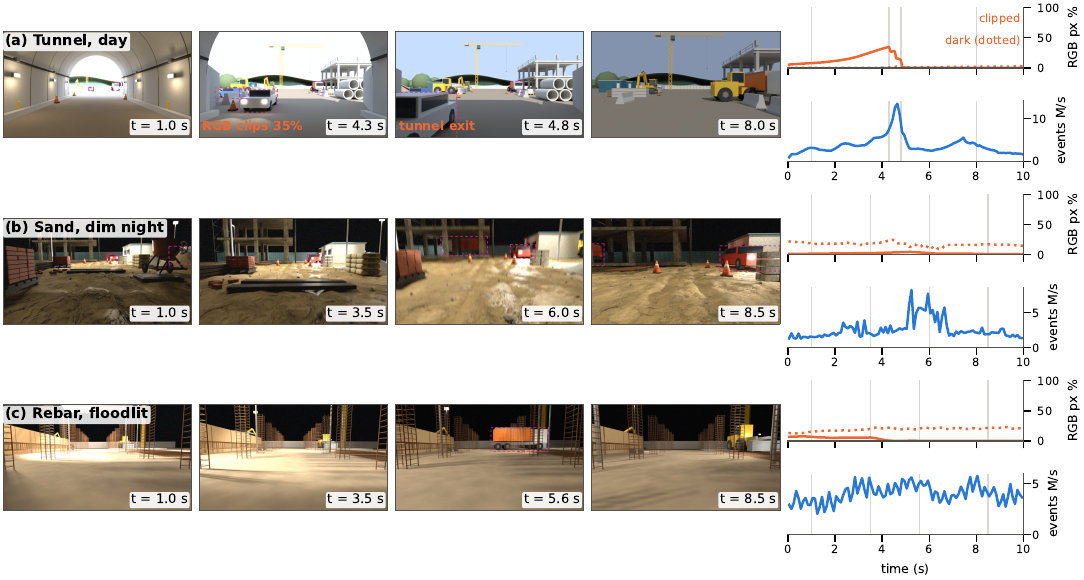}
\caption{The three simulated construction sites over their 10 s runs: (a) tunnel exit by day, (b) sandy yard at dim night, (c) rebar yard under floodlights. Left: simulated RGB frames at four timestamps (labelled objects dashed). Right: share of clipped ($>$250, solid) and dark ($<$30, dotted) RGB pixels, and the simulated event rate; grey lines mark the shown timestamps.}
\label{fig:simtl}
\end{figure*}

PEOD is a driving dataset, so we rendered three construction sites in Blender (Cycles) to test whether the conclusions carry over. A ground-vehicle camera drives through each site for 10 s while a car, a dump truck, a wheel loader and an excavator move or work nearby, and a car stands parked; no people are included. Figure~\ref{fig:simtl} shows each scenario over time.

\textbf{Tunnel.} The camera (1.7 m high, 4.25 m/s) drives out of a construction tunnel lit by work lamps into a sunlit yard, meeting an oncoming car with its headlights on; a wheel loader crosses the yard from 3 s and a dump truck approaches from 4 s. As the bright portal fills the view, the simulated RGB camera clips 35\% of its pixels at 4.3 s, and its auto-exposure recovers only after the exit, around 4.7 s, when the event rate peaks at about 13 M events/s (Figure~\ref{fig:simtl}a). Variants: day and low sun.

\textbf{Sandy yard.} The camera (0.6 m high) drives slowly along an uneven sandy lane past a building with scaffolding and a site office; a dump truck crosses the site between 1 and 9 s, a wheel loader and a slow car approach, and an excavator slews back and forth (45$^\circ$, 7 s period). Variants: day, floodlit night (four floodlight towers) and dim night, in which the floodlights run at one tenth of their power. In the dim variant the RGB camera's gain stays at its 40$\times$ limit for most of the run (Figure~\ref{fig:simtl}b).

\textbf{Rebar yard.} The camera (0.6 m high, 1.2 m/s) drives between rebar column cages, mats, formwork and a purlin canopy; a dump truck crosses the frame between 1 and 9 s, a car approaches and a wheel loader moves forward from 2 s. The dense cage texture produces a high, steady event rate (median 4.0 M events/s, Figure~\ref{fig:simtl}c). Variants: day with striped canopy shadows and floodlit night (three towers).

Frames are rendered as linear radiance at 120 fps and 1280$\times$720. Events are simulated on log luminance (contrast threshold 0.25 $\pm$ 0.03, refractory period 0.5 ms, background noise), after spatial and temporal smoothing that suppresses render noise (2.6--5.9 M events/s). The RGB camera is simulated at 30 fps with lagging auto-exposure (time constant 0.5 s), a gain limit of 40$\times$, exposure-time blur, sensor noise and 8-bit clipping. Boxes come from projected 3D bounds with ray-cast visibility; cars and dump trucks are labelled as car and truck, while the loader and excavator have no PEOD class and are left unlabelled. All detectors are applied zero-shot, without fine-tuning.

\section{Results}

\subsection{Events vs RGB by lighting condition}

With the RGB detector trained on PEOD, event detection still leads in low light, while bright scenes are on par (Table~\ref{tab:1}).

\begin{table}[!htb]
\centering
\caption{AP (\%) on PEOD test by lighting subset, confidence $\geq$ 0.1 (z-s: zero-shot; tr.: trained on PEOD).}
\label{tab:1}
\small\setlength{\tabcolsep}{2.5pt}
\begin{tabular}{lccccc}
\hline
\textbf{Subset} & \textbf{Frames} & \textbf{RGB z-s} & \textbf{RGB tr.} & \textbf{SMamba} & \textbf{EvRT} \\ \hline
Low light & 8,161 & 9.0 & 21.3 & 19.4 & \textbf{27.2} \\
Overexposure & 681 & \textbf{37.8} & 23.8 & 17.8 & 23.5 \\
Well-lit & 2,817 & 37.8 & 39.7 & 30.7 & \textbf{42.0} \\
All & 11,659 & 14.6 & 27.1 & 22.4 & \textbf{32.2} \\
\hline
\end{tabular}
\end{table}

Training on PEOD raises RGB from 14.6 to 27.1 AP, but the best event detector still leads by 5.9 AP in low light. SMamba, without pre-training, falls slightly below trained RGB in low light. Fine-tuning lowers RGB on overexposed frames (37.8 $\rightarrow$ 23.8), possibly because such frames are rare in training. YOLO's confidences are lower than the event detectors', so results depend on the threshold: at confidence $\geq$ 0, trained RGB reaches 31.3 overall and 44.9 on well-lit frames (EvRT-DETR 42.4), while EvRT-DETR still leads in low light (27.7 vs 24.5). The low-light lead holds across scenes: in a bootstrap over test sequences EvRT-DETR beats trained RGB in every resample (95\% interval of the difference 3.5 to 10.8 AP), whereas the well-lit difference is not significant ($-$3.1 to 9.8).

At object level, EvRT-DETR finds 73\% of the 37,790 low-light objects and trained RGB 57\%; 6,669 low-light objects are found only by events, twelve times more than the 542 found only by RGB (Table~\ref{tab:2}). With the zero-shot RGB detector the imbalance is larger still: about half of the low-light objects are found only by SMamba, 20 times more than by RGB alone. Our overexposure subset is mostly bright daytime scenes with saturated sky, lamps or tunnel exits (Figure~\ref{fig:dataset}b), where objects stay visible to RGB; it does not capture a camera blinded by direct sun.
 
Workers are the objects that matter most on site, and PEOD's person class is the closest proxy (10,477 test objects, 71\% of them in low light; Table~\ref{tab:person}). In low light EvRT-DETR leads trained RGB in person AP by 4.6 points (AP$_{50}$ 66.8 vs 48.5) and finds 69\% of persons against 38\%. For small persons (below 32$^2$ px, the size of distant workers) it finds 30\% against 3\%, and 2,478 low-light persons are found only by events, 17 times the 143 found only by RGB. RGB's under-confident scores make these object-level rates partly threshold-dependent, but AP$_{50}$ shows the same ordering. In well-lit and overexposed frames RGB is better for persons (AP 23.0 vs 21.4 and 18.4 vs 9.7). Fusion is best in every subset: it finds 61\% of well-lit persons, against 50\% for EvRT-DETR and 47\% for RGB. Small persons remain the weak point for every detector.
 
\begin{table}[!htb]
\centering
\caption{Person class: AP (\%) by lighting subset (confidence $\geq$ 0.1) and share of low-light persons found (confidence $\geq$ 0.3, IoU $\geq$ 0.5), all sizes and small ($<$32$^2$ px).}
\label{tab:person}
\small\setlength{\tabcolsep}{3pt}
\begin{tabular}{lccccc}
\hline
\textbf{Model} & \textbf{All} & \textbf{LL} & \textbf{OE} & \textbf{WL} & \textbf{Found LL (small)} \\ \hline
RGB trained & 22.6 & 23.0 & 18.4 & 23.0 & 38\% (3\%) \\
SMamba & 18.8 & 20.0 & 5.0 & 16.9 & 53\% (8\%) \\
EvRT-DETR & 25.5 & 27.6 & 9.7 & 21.4 & 69\% (30\%) \\
Fused & \textbf{27.5} & \textbf{28.4} & \textbf{18.9} & \textbf{26.1} & \textbf{70\% (30\%)} \\
\hline
\end{tabular}
\end{table}

\begin{table}[!htb]
\centering
\caption{Ground-truth objects found by EvRT-DETR (events) and/or trained RGB (confidence $\geq$ 0.3, IoU $\geq$ 0.5), and by their late fusion.}
\label{tab:2}
\small\setlength{\tabcolsep}{2.5pt}
\begin{tabular}{lccccc}
\hline
\textbf{Subset} & \textbf{Objects} & \textbf{Ev. only} & \textbf{RGB only} & \textbf{Both} & \textbf{Fused} \\ \hline
Low light & 37,790 & \textbf{6,669} & 542 & 21,086 & 28,102 \\
Overexposure & 2,334 & \textbf{262} & 65 & 1,509 & 1,827 \\
Well-lit & 12,967 & \textbf{1,639} & 531 & 8,201 & 10,321 \\
\hline
\end{tabular}
\end{table}

\subsection{Fusing RGB and events}

Late fusion of trained RGB and EvRT-DETR detections is the best model in every lighting subset (Table~\ref{tab:3}), without any training.

\begin{table}[!htb]
\centering
\caption{AP (\%) of single sensors and their late fusion, confidence $\geq$ 0.1.}
\label{tab:3}
\begin{tabular}{lccc}
\hline
\textbf{Subset} & \textbf{RGB} & \textbf{EvRT} & \textbf{Fused} \\ \hline
Low light & 21.3 & 27.2 & \textbf{27.8} \\
Overexposure & 23.8 & 23.5 & \textbf{26.0} \\
Well-lit & 39.7 & 42.0 & \textbf{44.7} \\
All & 27.1 & 32.2 & \textbf{33.3} \\
\hline
\end{tabular}
\end{table}

\begin{figure*}[!htb]
\centering
\includegraphics[width=\textwidth]{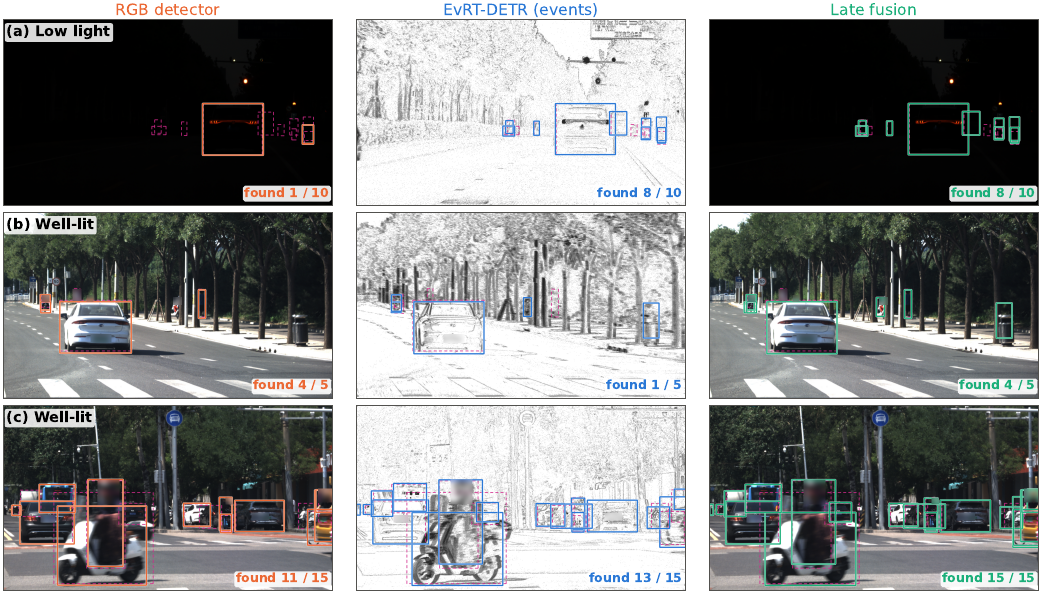}
\caption{How late fusion compensates for a failing sensor (PEOD test frames). Columns: trained RGB detector on the RGB frame, EvRT-DETR on the events of the same 33 ms, and their fusion (union + class-wise NMS) drawn on the RGB frame; solid boxes are detections with confidence $\geq$ 0.3, dashed boxes ground truth. (a) Low light, sequence 17: RGB fails, events succeed. (b) Daylight, sequence 16: events miss, RGB succeeds. (c) Daylight, sequence 13: the sensors find different objects. RGB is shown as recorded (not brightened).}
\label{fig:fusion}
\end{figure*}

Figure~\ref{fig:fusion} shows how fusion covers the failure of either sensor. In a dark street (a) RGB finds one of ten objects, the large car lit by its own lamps, whereas EvRT-DETR finds eight; the fused output keeps all eight. In daylight (b) the event stream is dominated by the texture of roadside trees and EvRT-DETR finds one of five objects, while RGB finds four; fusion again keeps the better sensor's boxes. Most often (c) each sensor finds objects the other misses, and the union finds all 15. Across the test set, fusion finds 76\% of all objects at the object-level threshold, against 74\% for EvRT-DETR and 60\% for RGB (Table~\ref{tab:2}).

Gains in AP are largest where RGB is informative (well-lit +2.7, overexposure +2.2) and small in low light (+0.6), where RGB adds few objects that events miss. The ranking holds at confidence $\geq$ 0 (fused 33.9, EvRT-DETR 32.6, RGB 31.3). Small-object AP depends on the threshold: at confidence $\geq$ 0.1 fusion doubles RGB's (5.1 vs 2.5), but at $\geq$ 0 it falls below RGB (5.2 vs 8.6), because EvRT-DETR outputs 17 times more boxes and its low-confidence boxes displace RGB's small detections. Fusion beats EvRT-DETR in 98\% of the sequence resamples (+1.1 AP, 95\% interval 0.0 to 2.4). The two detectors' confidences are miscalibrated in opposite directions: 78\% of RGB detections above 0.1 are correct against a mean confidence of 0.62, whereas only 22\% of EvRT-DETR's are correct against 0.31. EvRT-DETR's overconfident boxes can therefore suppress correct RGB boxes in NMS. Platt scaling removes most of this error (expected calibration error 0.156 $\rightarrow$ 0.032 for RGB, 0.108 $\rightarrow$ 0.004 for EvRT-DETR), but fusing calibrated confidences gains only 0.2 AP (33.5), mostly in overexposed (+0.7) and well-lit frames (+0.3), and calibrating per lighting subset adds nothing (33.4). Weighting the event detector per lighting subset did not help either (32.4): the chosen weights varied between folds, although events always kept near-full weight in low light. Plain late fusion is thus already close to what score-level fusion can reach, and a better combination will need feature-level fusion.

\subsection{Brightness versus motion}

\begin{figure*}[!htb]
\centering
\includegraphics[width=\textwidth]{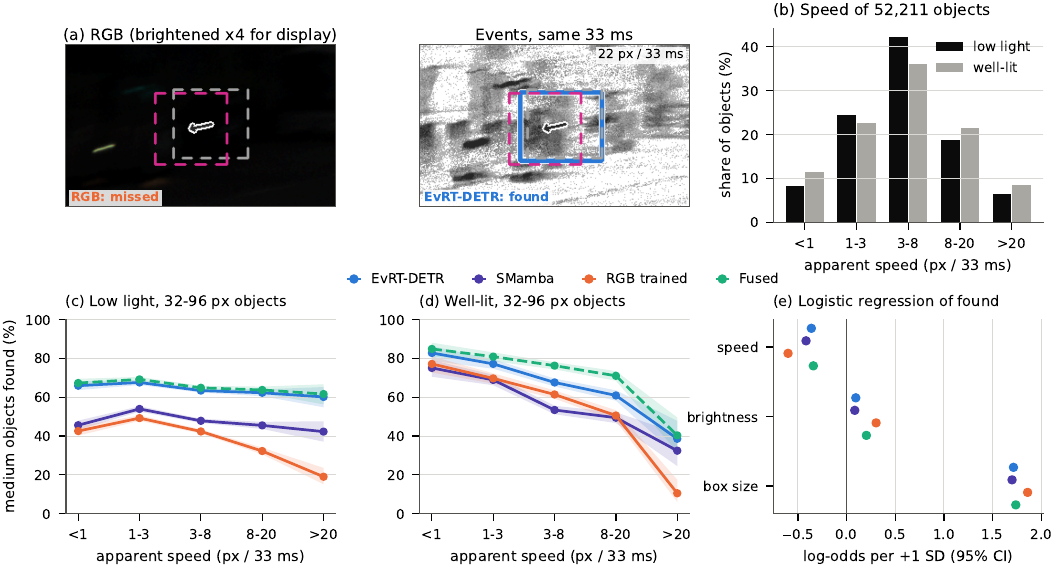}
\caption{Brightness versus motion on PEOD test. (a) How apparent speed is measured: a car in low light moving 22 px per 33 ms; grey dashed box: previous labelled frame, magenta dashed box: current frame, arrow: displacement; RGB missed it, EvRT-DETR found it (solid blue). (b) Share of objects per speed bin. (c, d) Medium-size objects found by speed in low light and well-lit frames, with Wilson 95\% bands. (e) Standardised logistic-regression coefficients of \emph{found} with 95\% intervals, 52,211 objects.}
\label{fig:motion}
\end{figure*}

For event detectors, brightness has a small effect on whether an object is found, while apparent speed has a larger one (Figure~\ref{fig:motion}e). Box size dominates for every detector (+1.7 to +1.9 per standard deviation). A one-standard-deviation increase in speed lowers the log-odds of detection by 0.36 for EvRT-DETR and 0.42 for SMamba but by 0.60 for trained RGB, and brightness matters about three times more for RGB (+0.30) than for the event detectors (+0.08 to +0.09). All 95\% intervals are narrower than $\pm$0.04.

The event advantage grows with speed (Figure~\ref{fig:motion}c,d). For medium-size objects in low light, EvRT-DETR leads trained RGB by 23 points for nearly static objects (66\% vs 43\%) and by 41 points above 20 px per frame (60\% vs 19\%); in well-lit frames the lead grows from 6 to 28 points. EvRT-DETR's detection rate in low light is almost flat across speeds, whereas RGB's falls steadily, consistent with motion blur under long exposures. Darkness costs EvRT-DETR about 17 points for slow objects (66\% vs 83\%) but not for fast ones. Static objects are not invisible to event cameras: EvRT-DETR still finds 66--89\% of objects moving less than 1 px per frame, helped by camera motion. For event detection, motion matters more than brightness, and fast motion is where events help most. Low-light and well-lit frames come from different scenes, so these comparisons are observational; the fastest well-lit bin holds only 114 medium-size objects, hence its wide intervals.

\begin{figure}[!htb]
\centering
\includegraphics[width=\columnwidth]{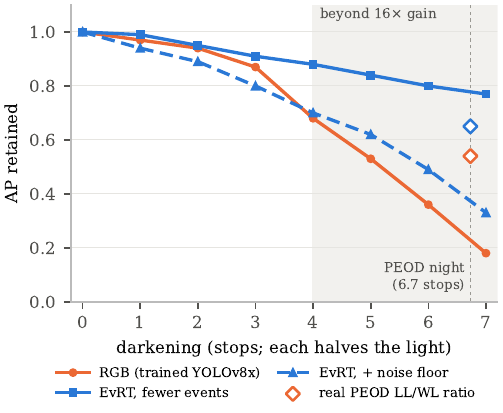}
\caption{Controlled darkening of the six brighter test sequences: AP retained relative to 0 stops (full-input AP: RGB 40.3, EvRT-DETR 42.7). EvRT-DETR's input is thinned to the event share measured at each light level, alone (solid) or with a constant noise floor (dashed). Shaded: beyond the RGB camera's 16$\times$ gain limit. Diamonds: real PEOD ratio of low-light to well-lit AP.}
\label{fig:dark}
\end{figure}

Darkening identical scenes reproduces the pattern (Figure~\ref{fig:dark}). Up to 3 stops, while auto-gain compensates, trained RGB keeps 87\% of its AP and EvRT-DETR 80--91\%. Beyond the gain limit RGB collapses, to 36\% at 6 stops and 18\% at 7, whereas EvRT-DETR keeps 77--80\% when only events are lost and 33--49\% when the noise floor stays constant; at PEOD's night level both event models stay above RGB. The real ratios of low-light to well-lit AP in PEOD (EvRT-DETR 0.65, RGB 0.54) lie between our two event models and above our RGB model, probably because the real camera has more exposure headroom than our 16$\times$ limit and night scenes contain lamps and headlights. The controlled darkening therefore confirms the direction of the effect rather than its exact size.

\subsection{Why EvRT-DETR beats SMamba}

\begin{figure*}[!htb]
\centering
\includegraphics[width=\textwidth]{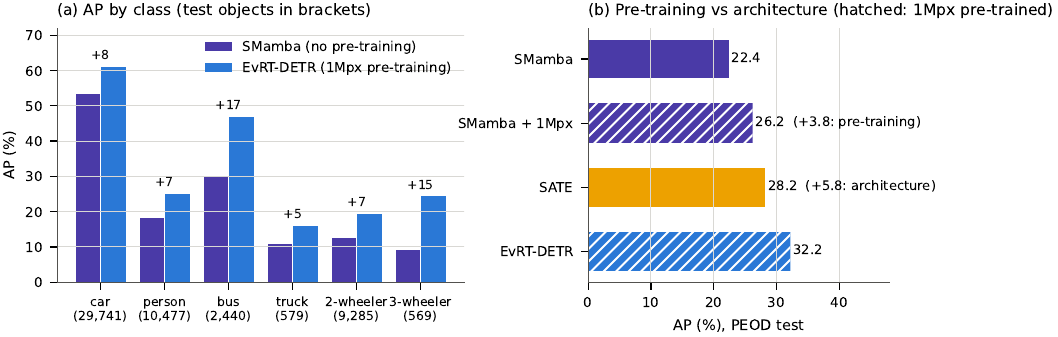}
\caption{Why EvRT-DETR beats SMamba. (a) AP per class on PEOD test (number of test objects in brackets); labels give EvRT-DETR's lead. (b) Controlled runs that separate 1Mpx pre-training (hatched) from architecture. SMamba + 1Mpx is evaluated at half resolution.}
\label{fig:evrtsm}
\end{figure*}

EvRT-DETR is 9.8 AP better than SMamba overall and better in every lighting subset (+7.8 low light, +11.3 well-lit, +5.7 overexposure). It does not close the illumination gap: the ratio of low-light to well-lit AP is 0.65 for EvRT-DETR and 0.63 for SMamba. It is better for every class, and gains are largest for the rare classes (Figure~\ref{fig:evrtsm}a): bus +17 AP and 3-wheeler +15 AP, against +8 for cars, which make up 56\% of the test objects. This pattern points to data scale. Its stage-2 temporal fusion adds less than 1 AP.

Two controlled runs separate pre-training from architecture (Figure~\ref{fig:evrtsm}b). Fine-tuning SMamba from its 1Mpx checkpoint raises AP from 22.4 to 26.2 (AP$_{50}$ 38.3 $\rightarrow$ 47.8). SATE, trained from scratch on PEOD for half as many steps as SMamba, reaches 28.2 AP (AP$_{50}$ 53.2) and raises small-object AP from 1.7 to 9.2, above EvRT-DETR's 5.0. Pre-training accounts for about 4 AP and architecture for about 6 AP of the gap, and SATE's early temporal encoding helps the small objects that matter for distant workers.

Pre-training and early temporal encoding both help most in the dark (Table~\ref{tab:light}). Relative to SMamba, 1Mpx pre-training adds 5.6 AP in low light but only 1.6 in well-lit frames, and SATE adds 5.4 and 2.4, so the ratio of low-light to well-lit AP rises from 0.63 to 0.75--0.77. EvRT-DETR's further lead over SATE comes mostly from well-lit frames (+8.9, against +2.4 in low light), so its ratio falls back to 0.65. For night work, where low-light accuracy matters most, SATE, trained on PEOD alone without pre-training, comes within 2.4 AP of EvRT-DETR.

\begin{table}[!htb]
\centering
\caption{AP (\%) of the event detectors by lighting subset, confidence $\geq$ 0.1. LL: low light, OE: overexposure, WL: well-lit; LL/WL: ratio of low-light to well-lit AP.}
\label{tab:light}
\small\setlength{\tabcolsep}{3pt}
\begin{tabular}{lccccc}
\hline
\textbf{Model} & \textbf{All} & \textbf{LL} & \textbf{OE} & \textbf{WL} & \textbf{LL/WL} \\ \hline
SMamba & 22.4 & 19.4 & 17.8 & 30.7 & 0.63 \\
SMamba + 1Mpx & 26.2 & 25.0 & 21.1 & 32.3 & 0.77 \\
SATE & 28.2 & 24.8 & 20.8 & 33.1 & 0.75 \\
EvRT-DETR & 32.2 & 27.2 & 23.5 & 42.0 & 0.65 \\
\hline
\end{tabular}
\end{table}

\subsection{Token pruning}

\begin{figure*}[!htb]
\centering
\includegraphics[width=\textwidth]{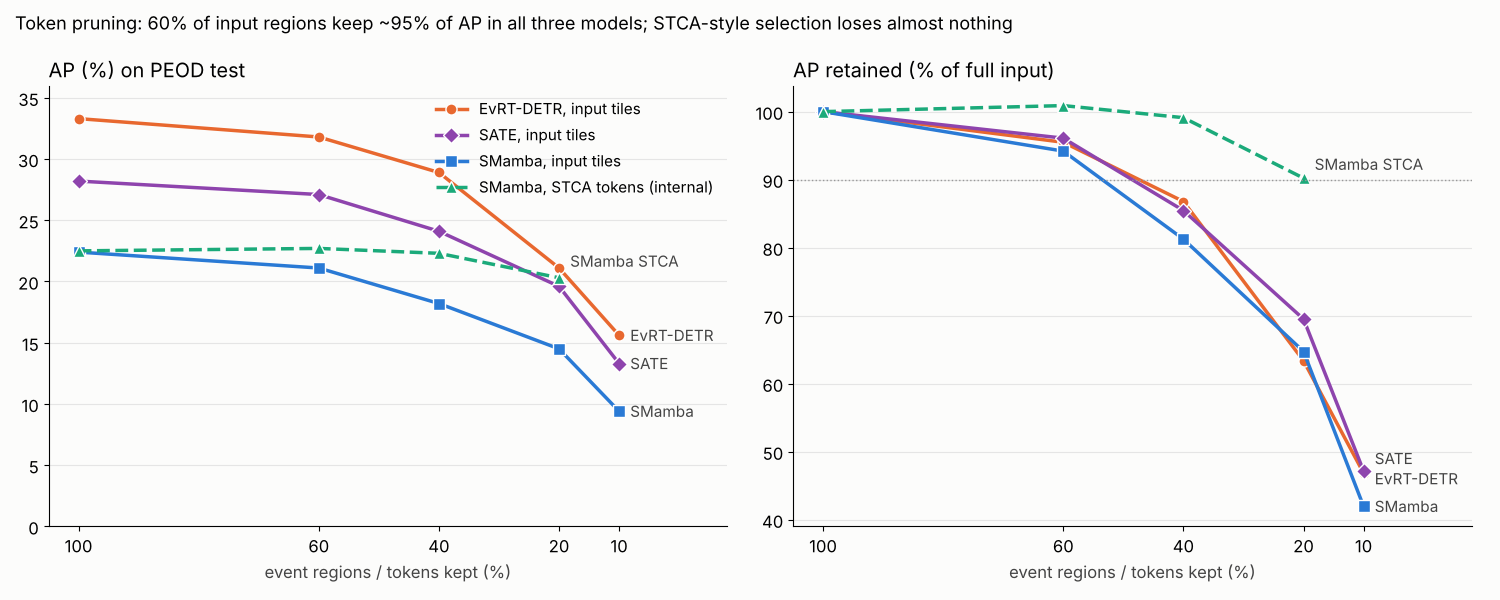}\\[2pt]
\includegraphics[width=\textwidth]{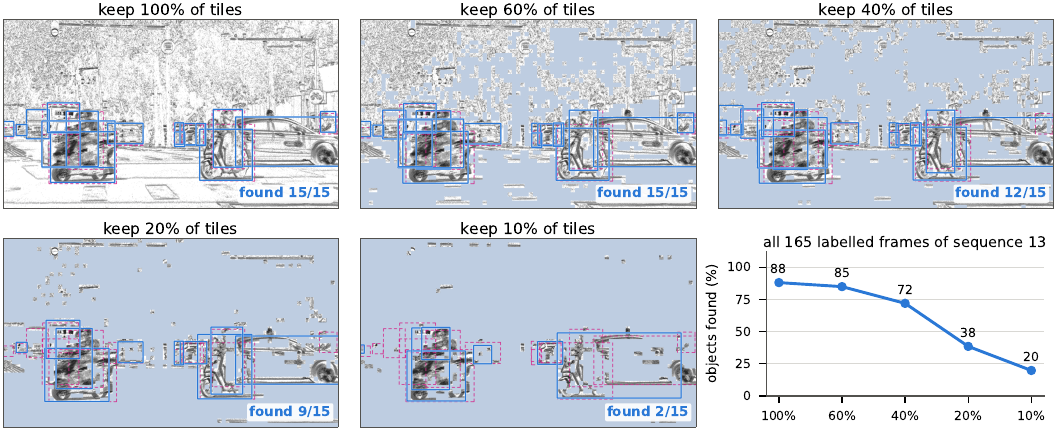}
\caption{Input-tile pruning (top-k tiles by event count). Top: AP retention on PEOD test; STCA = SMamba's internal token selection, approximate keep ratio. Full-input AP in the pruning runs (same hook, all tiles kept): EvRT-DETR 33.3, SATE 28.2, SMamba 22.4; retention is relative to these values. Bottom: EvRT-DETR on one well-lit test frame (sequence 13) at decreasing keep ratios; removed tiles shaded (mask recomputed at full resolution for display), detections with confidence $\geq$ 0.3 solid, ground truth dashed; the last panel gives the share of objects found over the whole sequence.}
\label{fig:prune}
\end{figure*}

All three detectors keep about 95\% of their AP when only 60\% of input regions are kept, and their retention curves nearly coincide (Figure~\ref{fig:prune}, top). Redundancy is therefore a property of the event data rather than of any one architecture. Below 40\% the curves bend down and AP collapses: keeping 10\% of the tiles leaves EvRT-DETR with 15.6 AP and SMamba with 9.4. The bottom of Figure~\ref{fig:prune} shows where the collapse happens on a busy daylight frame. With 60\% of the tiles EvRT-DETR still finds all 15 objects, because the removed tiles are mostly flat road and sky; at 40\% it finds 12, and below that the removed tiles reach the objects themselves, leaving 9 objects at 20\% and 2 at 10\%. Over all 165 labelled frames of this sequence the share of objects found falls from 88\% with the full input to 85\%, 72\%, 38\% and 20\%. SMamba's internal STCA selection keeps 99\% of AP at 40\% of tokens, so how tokens are chosen matters more than how many.

\begin{figure*}[!htb]
\centering
\includegraphics[width=\textwidth]{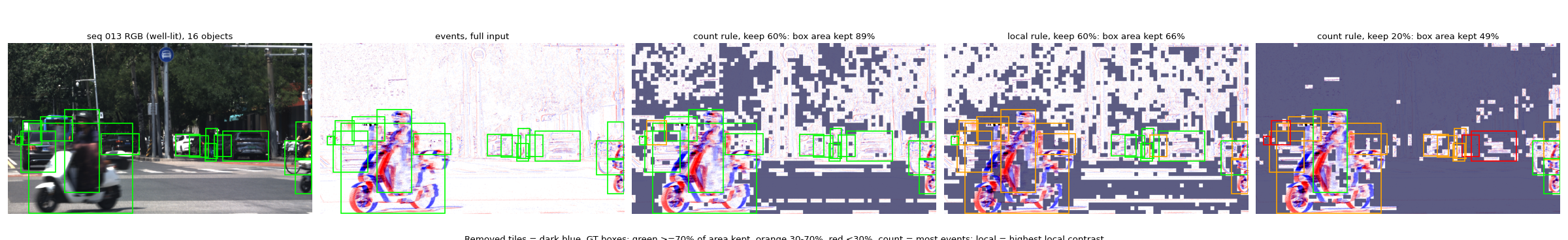}
\caption{Removed tiles in dark blue; boxes green when $\geq$ 70\% of their area is kept, orange 30--70\%, red < 30\%. The count rule removes flat road and sky; the local rule removes the interior of large objects.}
\label{fig:5}
\end{figure*}

Small objects are the most fragile, but not for the obvious reason. Removing 10\% of regions barely affects SATE (AP 28.3; small-object AP 9.2 $\rightarrow$ 8.8), while at 60\% kept its small-object AP falls to 3.8 and small-object recall from 11.9 to 6.6. Yet the pruned input keeps 99\% of the area of small boxes and 93\% of a surrounding ring three times the box size (Table~\ref{tab:6}), more than for large objects. Losing the object's own events or its immediate context therefore does not explain the drop, and recall, not only precision, falls. Ranking tiles by local contrast (event count relative to the surrounding 7$\times$7 tiles) keeps less of small boxes (93\%) yet raises small-object AP to 10.6 and recall to 12.4, above the full input, at the cost of overall AP (23.8) because it removes the interior of large objects (Figure~\ref{fig:5}). Selection criteria thus trade small-object against large-object accuracy; understanding why small objects are lost is future work, and it matters on construction sites, where distant workers are small in the image.

\begin{table}[!htb]
\centering
\caption{Share (\%) of GT box area and of a surrounding ring (3$\times$ box) kept by pruning, 2,695 boxes on every 20th test frame.}
\label{tab:6}
\small\setlength{\tabcolsep}{3pt}
\begin{tabular}{lcccccc}
\hline
\textbf{Size} & \multicolumn{2}{c}{\textbf{Count, 60\%}} & \multicolumn{2}{c}{\textbf{Local, 60\%}} & \multicolumn{2}{c}{\textbf{Count, 20\%}} \\
& box & ring & box & ring & box & ring \\ \hline
Small & 99 & 93 & 93 & 67 & 88 & 54 \\
Medium & 97 & 80 & 84 & 60 & 74 & 36 \\
Large & 89 & 62 & 71 & 59 & 55 & 22 \\
\hline
\end{tabular}
\end{table}

\subsection{Inside the detectors}

\begin{figure*}[!htb]
\centering
\includegraphics[width=\textwidth]{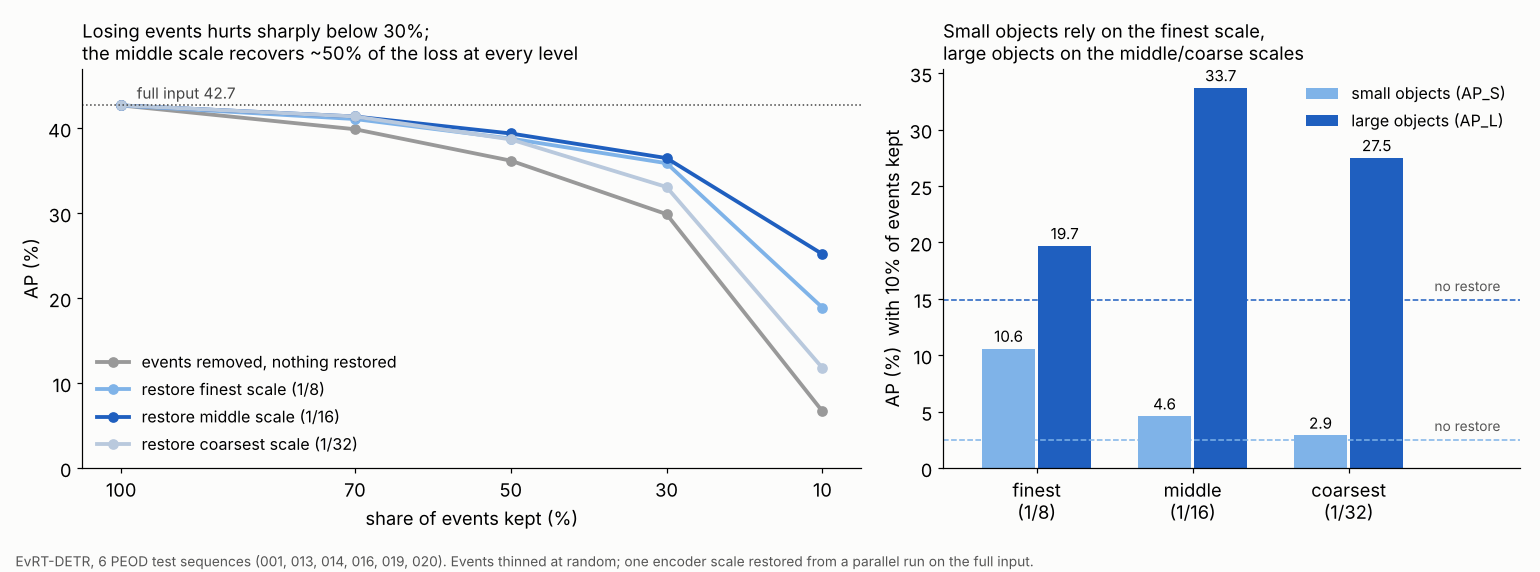}
\caption{EvRT-DETR on six brighter test sequences (full-input AP 42.7). Events thinned at random; one scale restored from a parallel run on the full input.}
\label{fig:6}
\end{figure*}

\begin{figure*}[!htb]
\centering
\includegraphics[width=\textwidth]{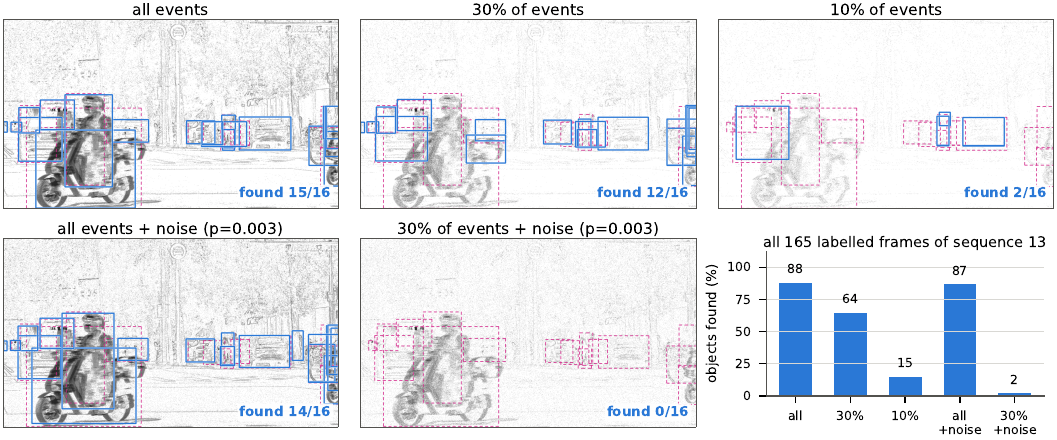}
\caption{EvRT-DETR on one well-lit test frame (sequence 13) as events are thinned at random and uniform noise is added (p: noise probability per pixel and time bin), mimicking a darkening scene. Panels show a re-sampled thinning for display, on the same grey scale as the full input; solid: detections (confidence $\geq$ 0.3), dashed: ground truth. Last panel: share of objects found over the whole sequence.}
\label{fig:losscases}
\end{figure*}

Losing events hurts mainly at the 1/16-resolution features, and the damage escalates when few signal events meet noise (Figures~\ref{fig:6} and~\ref{fig:losscases}, Table~\ref{tab:7}).

\textbf{Activation patching.} EvRT-DETR loses little until half of the events are gone (36.2 AP at 50\%) and collapses below 30\% (6.7 AP at 10\%). Restoring the 1/16 scale recovers about half of the loss at every level, more than the 1/8 or 1/32 scale. SMamba shows the same pattern: with 30\% of events kept, restoring its 1/16 stage recovers 54\% of the loss. Scales specialise by object size: with 10\% of events kept, restoring the finest scale lifts small-object AP from 2.5 to 10.6, while restoring the 1/16 scale lifts large-object AP from 14.9 to 33.7.

\begin{table}[!htb]
\centering
\caption{EvRT-DETR AP under event loss and random noise events (probability p per pixel and time bin; real event density is about 1\% of pixel-bins).}
\label{tab:7}
\small\setlength{\tabcolsep}{3pt}
\begin{tabular}{lcccc}
\hline
\textbf{Kept} & \textbf{No noise} & \textbf{p=0.001} & \textbf{p=0.003} & \textbf{p=0.01} \\ \hline
100\% & 42.7 & 38.4 & 32.8 & 20.0 \\
30\% & 29.9 & 13.9 & 3.7 & 0.3 \\
\hline
\end{tabular}
\end{table}

\textbf{Noise.} Noise and event loss are strongly super-additive: keeping 30\% of events costs 12.8 AP and light noise (p = 0.001, about 10\% extra events) costs 4.3, but together they cost 28.8. Figure~\ref{fig:losscases} shows the same interaction on a daylight frame: EvRT-DETR finds 15 of 16 objects with all events, 12 with 30\% of them and 14 with added noise (p = 0.003), but none when both are combined. Over the 165 labelled frames of that sequence, the share of objects found falls from 88\% to 64\% with event loss and stays at 87\% with noise alone, but drops to 2\% with both. A background-activity filter recovers part of the damage when events are plentiful: with all events and noise at p = 0.003 it restores AP from 32.8 to 38.6, more than half of the loss, and it costs nothing on clean input (42.7). When few events meet noise it helps little: with 30\% of events AP rises from 13.9 to 17.9 at p = 0.001 and from 3.7 to 7.6 at p = 0.003, far below the 29.9 reached without noise. The filter keeps 85--98\% of active cells, probably because real events cluster along edges and most noise cells have an active neighbour. A dark scene produces exactly this combination, so the signal-to-noise ratio, not brightness itself, is what breaks event detection. Our noise is uniform and synthetic; real sensor noise (shot noise, hot pixels, lamp flicker) is structured.

\textbf{Temporal memory and probes.} Resetting EvRT-DETR's ConvLSTM memory at every frame costs 1.1 AP on 11 dark sequences and 1.4 AP on 8 bright ones, so temporal memory does not specifically compensate for darkness. Linear probes decode low light from SMamba features with 0.88--0.90 balanced accuracy, against 0.67 from the event count alone: the detector knows it is dark, but, as Section~3.3 showed, this barely changes whether it finds objects. Steering low-light features towards the well-lit mean did not improve AP.

\subsection{Zero-shot transfer to simulated sites}

Absolute AP on the simulated sites is low for both sensors (Table~\ref{tab:sim}) because rendered vehicles and plant differ from PEOD's road vehicles; the trained RGB detector, for example, never recognises the rendered dump truck. We read the table across lighting conditions within a site. Dimming the floodlights of the sandy yard tenfold cuts RGB from 11.1 to 0.3 AP, while EvRT-DETR keeps most of its AP$_{50}$ (24.1 to 20.8). In the rebar yard RGB fails in both lighting conditions: the cages produce one false box covering most of the frame. In the tunnel, RGB has the higher AP over the whole run.

\begin{table}[!htb]
\centering
\caption{Zero-shot AP / AP$_{50}$ (\%) on the simulated sites.}
\label{tab:sim}
\small\setlength{\tabcolsep}{4pt}
\begin{tabular}{llcc}
\hline
\textbf{Site} & \textbf{Lighting} & \textbf{RGB} & \textbf{EvRT} \\ \hline
Tunnel & day & 13.7 / 18.8 & 5.5 / 9.5 \\
Tunnel & low sun & 18.5 / 23.7 & 8.3 / 13.6 \\
Rebar & day & 0.0 / 0.0 & 17.7 / 34.4 \\
Rebar & floodlit night & 0.0 / 0.0 & 18.5 / 35.3 \\
Sand & day & 10.4 / 18.3 & 4.1 / 11.9 \\
Sand & floodlit night & 11.1 / 16.8 & 6.8 / 24.1 \\
Sand & dim night & 0.3 / 0.5 & 4.9 / 20.8 \\
\hline
\end{tabular}
\end{table}

\begin{figure*}[!htb]
\centering
\includegraphics[width=\textwidth]{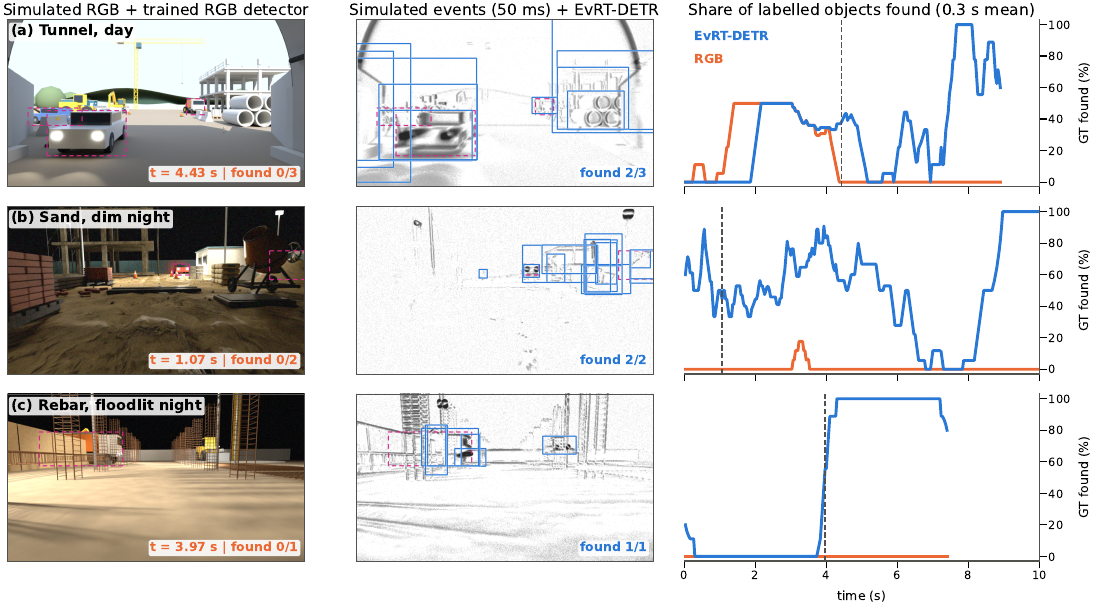}
\caption{Zero-shot detection on the simulated sites, at the timestamp where events succeed and RGB fails most clearly. Left: simulated RGB with trained-RGB detections (orange, confidence $\geq$ 0.1); middle: 50 ms of simulated events with EvRT-DETR detections (blue, confidence $\geq$ 0.15); dashed: labelled objects. Right: share of labelled objects found (IoU $\geq$ 0.5, 0.3 s moving mean) over the run; the dashed line marks the shown timestamp.}
\label{fig:simres}
\end{figure*}

Following the runs over time shows where each sensor succeeds (Figure~\ref{fig:simres}). In the tunnel, RGB and EvRT-DETR find similar shares of the labelled vehicles while the camera is inside. When the portal saturates the RGB image (4.3 s, 35\% of pixels clipped) RGB finds none of the three labelled vehicles, whereas EvRT-DETR finds two (Figure~\ref{fig:simres}a); after the exit RGB keeps missing, mainly because it never recognises the rendered dump truck, while EvRT-DETR finds most labelled objects. In the dim sandy yard RGB finds 0.5\% of the labelled objects over the whole run, EvRT-DETR 47.5\% (b). In the floodlit rebar yard RGB finds none, while EvRT-DETR localises the dump truck through the cages whenever it is in view, 70\% of its labelled appearances (c). EvRT-DETR's confidences are low on rendered data, however, and it also fires on cages, scaffolding and parked plant, so its boxes need the low threshold used here (0.15).

\section{Implications for construction sites}

Event cameras pay off where RGB is starved of light: night shifts, dusk, basements and tunnels. In bright scenes a good RGB detector is as good or better. They should therefore complement RGB cameras, not replace them; our fusion results show that the combination beats either sensor alone in every lighting condition, and because the fusion works on final detections, it can be added to an existing RGB system without retraining it.

Two further findings shape deployment. First, motion matters more than brightness for events, and their advantage is largest for fast-moving objects, such as plant and vehicles passing close to the camera or cameras mounted on moving machines. Second, event detection fails when few signal events meet noise. On site this means avoiding flickering AC lighting and dense static texture (rebar meshes, scaffolding) that floods the sensor with background events, and using the sensor's built-in noise and anti-flicker filters.

\textbf{Recommended field recordings.} (i) Night and dusk work, recording the lamp type, since AC lighting flickers and adds noise events; (ii) dark-to-bright transitions when a vehicle-mounted camera leaves a basement or tunnel, where RGB auto-exposure lags; (iii) low sun, glare and reflective surfaces, the real overexposure that PEOD lacks; (iv) the same route at morning, noon, dusk and night, so lighting is the only variable; (v) moving workers and plant, with static cones as reference; (vi) co-mounted, synchronised RGB and event cameras for pixel-level comparison. Recordings with workers require consent, and faces must be blurred for publication.

\section{Conclusion}

Against an RGB detector trained on the same pixel-aligned frames, event cameras lead by about 6 AP in low light, which covers 70\% of PEOD's test frames and matters most for night-time site safety, while RGB is on par in bright scenes; fusing both is best in every condition. For event detectors, brightness itself has little effect: what matters is motion, where events degrade far less than RGB, and the signal-to-noise ratio, which collapses when few events meet noise. Inside the detectors, the 1/16-resolution features carry most of the damage from losing events, and small objects depend on the finest scale and are the most fragile under pruning. For persons, the closest proxy for workers, the event detector finds 69\% in low light against 38\% for RGB, and 30\% of small, distant persons against 3\%. Zero-shot tests on simulated construction sites show the same pattern under dim lighting, tunnel exits and dense rebar. These results support installing event cameras alongside RGB cameras on construction sites, where night work, fast-moving plant and mobile platforms are common.

\bibliography{ISARC}

\end{document}